\documentclass[tablecaption=bottom,wcp]{article} 
\usepackage{tikz}
\usepackage{amsmath}
\usepackage{amssymb}
\usepackage{booktabs}
\usepackage{comment}
\usepackage[utf8]{inputenc}
\usepackage[english]{babel}
\usepackage[top=1in, bottom=1in, left=1in, right=1in]{geometry}
\usepackage{natbib}
\DeclareMathOperator*{\argmax}{arg\,max}
\DeclareMathOperator*{\argmin}{arg\,min}

\title{Training conformal predictors}

\author{
	Nicolo Colombo
	 and Vladimir Vovk \\
	 \vspace{.5cm}
  \small{
	  \{nicolo.colombo,v.vovk\}@rhul.ac.uk}\\
	  \vspace{.5cm}
  \footnotesize{Department of Computer Science, 
  Royal Holloway University of London, 
  Egham, UK}
  }

\begin{document}

\maketitle
\begin{abstract}
  Efficiency criteria for conformal prediction, such as \emph{observed fuzziness}
  (i.e., the sum of p-values associated with false labels),
  are commonly used to \emph{evaluate} the performance of given conformal predictors.
  Here, we investigate whether it is possible to exploit efficiency criteria to \emph{learn} classifiers,
  both conformal predictors and point classifiers,
  by using such criteria as training objective functions.

  The proposed idea is implemented for the problem of binary classification of hand-written digits.
  By choosing a 1-dimensional model class (with one real-valued free parameter),
  we can solve the optimization problems through an (approximate) exhaustive search
  over (a discrete version of) the parameter space.
  Our empirical results suggest that conformal predictors
  trained by minimizing their observed fuzziness
  perform better than conformal predictors trained in the traditional way
  by minimizing the \emph{prediction error} of the corresponding point classifier.
  They also have reasonable a performance in terms of their prediction error on the test set.
\end{abstract}

\section{Introduction}


The standard approach to designing conformal predictors
is to start from an existing machine-learning algorithm
and turn it into a conformity measure (there may be more than one way of doing this).
The rest is automatic: once we have a conformity measure,
we can compute p-values, prediction sets, predictive distributions, etc.
In this approach conformal prediction plays the role of a superstructure over traditional machine learning.
The two parts of the resulting prediction algorithms,
the traditional machine-learning part and the conformal part,
are fairly autonomous and the interface between them is limited.

The standard approach has been fairly successful;
conformal predictors have been built on top of a wide variety of traditional algorithms,
including the Lasso \citep{Lei2019}, deep learning \citep{CCBender2019},
ridge regression, nearest neighbours, support vector machines, decision trees, and boosting
\citep[Sections 2.3, 3.1, 4.2]{Vovketal2005}.
However, the separation of the two parts may limit the power of this approach.

In this paper we propose blending the two parts of standard conformal prediction.
The idea is to use existing criteria of efficiency for conformal prediction,
such as those defined in \citet{vovk2016criteria}.
Instead of \emph{evaluating} the performance of given conformal predictors,
we propose to use those criteria for \emph{training} conformal predictors
by using such criteria as training objective functions.

We demonstrate the idea using binary classification of hand-written digits as example;
see Section~\ref{sec:methods}.
As criterion of efficiency we use \emph{observed fuzziness},
defined to be the sum of p-values associated with false labels.
This is one of the \emph{probabilistic} criteria of efficiency;
they are defined by \citet{vovk2016criteria},
who argue that such criteria are akin to proper loss functions in machine learning
and should be used in practice.
The advantages of observed fuzziness over the other probabilistic criteria
defined in \citet{vovk2016criteria}
is that it does not depend on the significance level (which makes it easier to use)
and does not include the noise created by the p-value for the true label
(which makes it more stable).

For a simple 1-dimensional model class (i.e., involving one real-valued free parameters)
we can solve the optimization problems used for training through an approximate exhaustive search
over a discrete version of the parameter space.
We compare two ways of training conformal predictors and point classifiers:
using observed fuzziness and, as in traditional machine learning, using prediction error.
In this context, the distinction between conformal predictors and point classifiers blurs:
the former can be used as latter
(by using the label with the largest p-value as point prediction)
and the latter, being defined in terms of a conformity measure,
can be extended to the former.
Our empirical results suggest, not surprisingly,
that classifiers trained by minimizing observed fuzziness
lead to a better observed fuzziness on the test set,
and classifiers trained by minimizing prediction error
lead to a better (but not overwhelmingly better) prediction error on the test set.

For computational efficiency,
in this paper we concentrate on split-conformal prediction.
Using full conformal prediction and other directions of further research
are discussed in Section~\ref{sec:conclusion}.

\section{Background}

The set of natural numbers is denoted $\mathbb{N}:=\{1,2,\dots\}$
(the positive integers).
If $a$ and $b$ are two disjoint bags, we let $a+b:=a\cup b$,
and we use the notation $a+b$ only when the two (or more) bags are disjoint.

\subsection{Observation space and data sets}

Let ${\cal X}$ be a nonempty measurable \emph{object space},
${\cal Y}$ a discrete and finite \emph{label space} of size $\lvert{\cal Y}\rvert\ge2$,
and ${\cal Z} = {\cal X} \times {\cal Y}$ the corresponding \emph{observation space}.
We will refer to elements of these sets as \emph{objects}, \emph{labels} and \emph{observations},
respectively.

A \emph{dataset} is a bag of elements of $\mathcal{Z}$,
where a \emph{bag} is a collection of elements (observations in this case) some of which may be identical
\citep[Section 2.2]{Vovketal2005}.
We will use the notation $\mathcal{Z}^{(*)}$ for the set of all datasets.
Let us fix two nonempty datasets,
the \emph{training set} $\mathcal{D}_{\text{train}}$
and the \emph{test set} $\mathcal{D}_{\text{test}}$.



In split-conformal prediction,
the training set is randomly split into two disjoint bags, a \emph{pre-training set} and a \emph{pre-test set},
which we will denote as
\begin{equation}\label{eq:split}
  \mathcal{D}_{\text{train}} = \mathcal{D}_{\text{pre-train}} + \mathcal{D}_{\text{pre-test}}.
\end{equation}
The pre-training set is often called the proper training set
and the pre-test set is often called the calibration set
\citep[Section 4.1]{Vovketal2005},
but our current terminology will be more convenient for this paper.

\subsection{Conformity scores and p-values}

A \emph{conformity measure} is a function $Q: \mathcal{Z} \times \mathcal{Z}^{(*)} \to \mathbb{R}$
mapping each observation $(x, y)\in\mathcal{Z}$ and dataset $\mathcal{D}$
(such as the pre-training set $\mathcal{D}_{\text{pre-train}}$)
to the corresponding \emph{conformity score} $Q((x,y),\mathcal{D}) \in \mathbb{R}$.

The \emph{p-value} associated with an observation $(x,y) \in \mathcal{Z}$
(such as $(x,y)\in\mathcal{D}_{\text{test}}$),
nonempty datasets $\mathcal{D}$ (such as $\mathcal{D}_{\text{pre-train}}$)
and $\mathcal{D}'$ (such as $\mathcal{D}_{\text{pre-test}}$),
and a conformity measure $Q$
is
\begin{equation*} 
  C(x,y,\mathcal{D},\mathcal{D}',Q)
  :=
  \frac{
    1 +
    \sum_{z\in\mathcal{D}'}
    \theta
    \left(
      Q((x,y),\mathcal{D})
      -
      Q(z, \mathcal{D})
    \right)}
  {1 + \lvert\mathcal{D}'\rvert},
\end{equation*}
where the step function $\theta$ is defined by
\begin{equation*}
  \theta(u)
  =
  \begin{cases}
    1 & \text{if $u \ge 0$}\\
    0 & \text{if $u < 0$}
  \end{cases},
  \quad
  u\in \mathbb{R}.
\end{equation*}
For $(x,y)\in\mathcal{D}_{\text{test}}$,
$C(x,y, \mathcal{D}_{\text{pre-train}},\mathcal{D}_{\text{pre-test}},Q)$
shows how likely $y$ is as the label of the test object $x$.

\subsection{Observed fuzziness OF}

The \emph{observed fuzziness} of a conformity measure $Q$
for nonempty datasets $\mathcal{D}$ (such as $\mathcal{D}_{\text{pre-train}}$),
$\mathcal{D}'$ (such as $\mathcal{D}_{\text{pre-test}}$)
and $\mathcal{D}''$ (such as $\mathcal{D}_{\text{test}}$)
is
\begin{equation}\label{eq:OF-test}
  {\text{OF}}(\mathcal{D},\mathcal{D}',\mathcal{D}'',Q)
  :=
  \frac{
    \sum_{(x,y)\in\mathcal{D}''}
    \sum_{y' \in \mathcal{Y}}
    (1 - \delta_{y,y'})
    C(x, y', {\cal D}, \mathcal{D}', Q)}
  {\lvert\mathcal{D}''\rvert},
\end{equation}
where $\delta_{y,y'}$ is defined by
\begin{equation*}
  \delta_{y,y'}
  =
  \begin{cases}
    1 & \text{if $y = y'$}\\
    0 & \text{otherwise}
  \end{cases},
  \quad
  y, y' \in \mathcal{Y}. 
\end{equation*}

Note that one may choose ${\cal D}'' := {\cal D}'$;
this may be useful at the stage of estimating a future observed fuzziness.
For example, we may use
${\text{OF}}(\mathcal{D}_{\text{pre-train}},\mathcal{D}_{\text{pre-test}},\mathcal{D}_{\text{pre-test}},Q)$
to estimate
${\text{OF}}(\mathcal{D}_{\text{pre-train}},\mathcal{D}_{\text{pre-test}},\mathcal{D}_{\text{test}},Q)$
before seeing the test set.
However, this introduces bias, since in this case each element
of $\mathcal{D}''$ is also present in $\mathcal{D}'$,
which is not expected to be the case for $\mathcal{D}'':=\mathcal{D}_{\text{test}}$.
Therefore, we also define the three-argument version
\begin{equation*}
  {\text{OF}}(\mathcal{D},\mathcal{D}',Q)
  :=
  \frac{
    \sum_{(x,y)\in\mathcal{D}'}
    \sum_{y' \in \mathcal{Y}}
    (1 - \delta_{y,y'})
    C(x, y', {\cal D}, \mathcal{D}'\setminus\{(x,y)\}, Q)}
  {\lvert\mathcal{D}'\rvert}
\end{equation*}
of \eqref{eq:OF-test}.

\subsection{Prediction error (PE)}

The point predictor $\phi:{\cal X}\times {\cal Z}^{(*)} \to {\cal Y}$
obtained from a conformity measure $Q$ and dataset ${\cal D}$ is defined as
\begin{equation*} 
  \phi(x, {\cal D}, Q) 
  \in
  \argmax_{y \in {\cal Y}}
  Q((x,y), {\cal D});
\end{equation*}
we will assume that the $\argmax$ is a singleton (which is the case in our experiments).
The \emph{prediction error} of a conformity measure $Q$
on nonempty datasets ${\cal D}$ (such as ${\cal D}_{\text{train}}$)
and ${\cal D}'$ (such as ${\cal D}_{\text{test}}$)
is defined as
\begin{equation*} 
  \text{PE}({\cal D}, {\cal D}', Q) 
  :=
  \frac{
    \sum_{(x,y)\in\mathcal{D}'}
    (1 - \delta_{y, y_*})}
  {\lvert{\cal D}'\rvert},
  \quad
  y_* := \phi(x, {\cal D}, Q).
\end{equation*}
Note that, in this case, it is possible to use the entire training set 
as an input of the conformity measure, 
i.e., to let $\mathcal{D} := {\cal D}_{\text{train}}$
instead of $\mathcal{D} := {\cal D}_{\text{pre-train}}$.

\section{Methods}
\label{sec:methods}


For our experiments,
in addition to the split \eqref{eq:split},
we consider a further split
\begin{equation*}
  {\cal D}_{\text{pre-train}}
  =
  {\cal D}_{\text{pre-pre-train}} + {\cal D}_{\text{pre-pre-test}}.
\end{equation*}
The overall split of the available data is
\begin{equation}\label{eq:full-split}
  {\cal D}_{\text{train}}
  +
  {\cal D}_{\text{test}}
  =
  {\cal D}_{\text{pre-pre-train}}
  +
  {\cal D}_{\text{pre-pre-test}}
  +
  {\cal D}_{\text{pre-test}}
  +
  {\cal D}_{\text{test}}.
\end{equation}

\subsection{Model}
\label{section methods model}

We consider the binary classification problem of recognizing given hand-written digits in the MNIST dataset.
We choose the conformity measure
\begin{equation*}
  Q_\rho((x, y), {\cal D})
  :=
  \frac{
    \sum_{(x',y')\in{\cal D}}
    \delta_{y,y'} \kappa_\rho(x, x')}
  {\sum_{(x',y')\in{\cal D}} \kappa_\rho(x, x')},
  \quad
  \kappa_\rho(x, x')
  := 
  e^{-\rho\left\| x - x'\right\|^2},
\end{equation*}
where $\rho \in {\cal R} \subseteq[0,\infty)$ is a free parameter. 
We use the datasets in the split~\eqref{eq:full-split}
to train four models,
$Q_*^{\text{PE}}$, $Q_*^{\text{pre-PE}}$, $Q_*^{\text{OF}}$ and $Q_*^{\text{pre-OF}}$,
defined by
\begin{align*}
  Q^{\text{PE}}_* &:= Q_{\rho_*}, &
  \rho_*
  &:=
  \argmin_{\rho \in {\cal R}}
  \text{PE}({\cal D}_{\text{pre-train}},{\cal D}_{\text{pre-test}},Q_\rho),\\
  Q^{\text{pre-PE}}_* &:= Q_{\rho_*}, &
  \rho_*
  &:=
  \argmin_{\rho \in {\cal R}}
  \text{PE}({\cal D}_{\text{pre-pre-train}},{\cal D}_{\text{pre-pre-test}},Q_\rho),\\
  Q^{\text{OF}}_* &:= Q_{\rho_*}, &
  \rho_*
  &:=
  \argmin_{\rho \in {\cal R}} 
  \text{OF}({\cal D}_{\text{pre-train}},{\cal D}_{\text{pre-test}},Q_\rho),\\
  Q^{\text{pre-OF}}_* &:= Q_{\rho_*}, &
  \rho_*
  &:=
  \argmin_{\rho \in {\cal R}} 
  \text{OF}({\cal D}_{\text{pre-pre-train}},{\cal D}_{\text{pre-pre-test}},Q_\rho).
\end{align*} 
We evaluate the optimized models through the following performance scores:
\begin{align*} 
  \text{PE-test/PE-train}
  &:=
  \text{PE}({\cal D}_{\text{train}}, {\cal D}_{\text{test}}, Q^{\text{PE}}_*),\\
  \text{PE-test/OF-train}
  &:=
  \text{PE}({\cal D}_{\text{train}}, {\cal D}_{\text{test}}, Q^{\text{OF}}_*),\\
  \text{OF-test/PE-train}
  &:=
  \text{OF}({\cal D}_{\text{pre-train}}, \mathcal{D}_{\text{pre-test}}, {\cal D}_{\text{test}}, Q^{\text{pre-PE}}_*),\\
  \text{OF-test/OF-train}
  &:=
  \text{OF}({\cal D}_{\text{pre-train}}, {\cal D}_{\text{pre-test}}, {\cal D}_{\text{test}}, Q^{\text{pre-OF}}_*).
\end{align*}
Notice that for OF-testing we use the ``pre-models'' $Q^{\text{pre-PF}}_*$ and $Q^{\text{pre-OF}}_*$.
This is because we want a valid conformal predictor;
therefore, we do not touch the pre-test set at the stage of choosing our model.
On the other hand, there is no need to worry about validity in the case of PE-testing;
it is not guaranteed anyway.

\subsection{Experiments}


\foreach \x in {0}
{

\begin{figure}
\begin{center}
	\includegraphics[trim=200 100 200 100, clip=1, width=.90\textwidth]{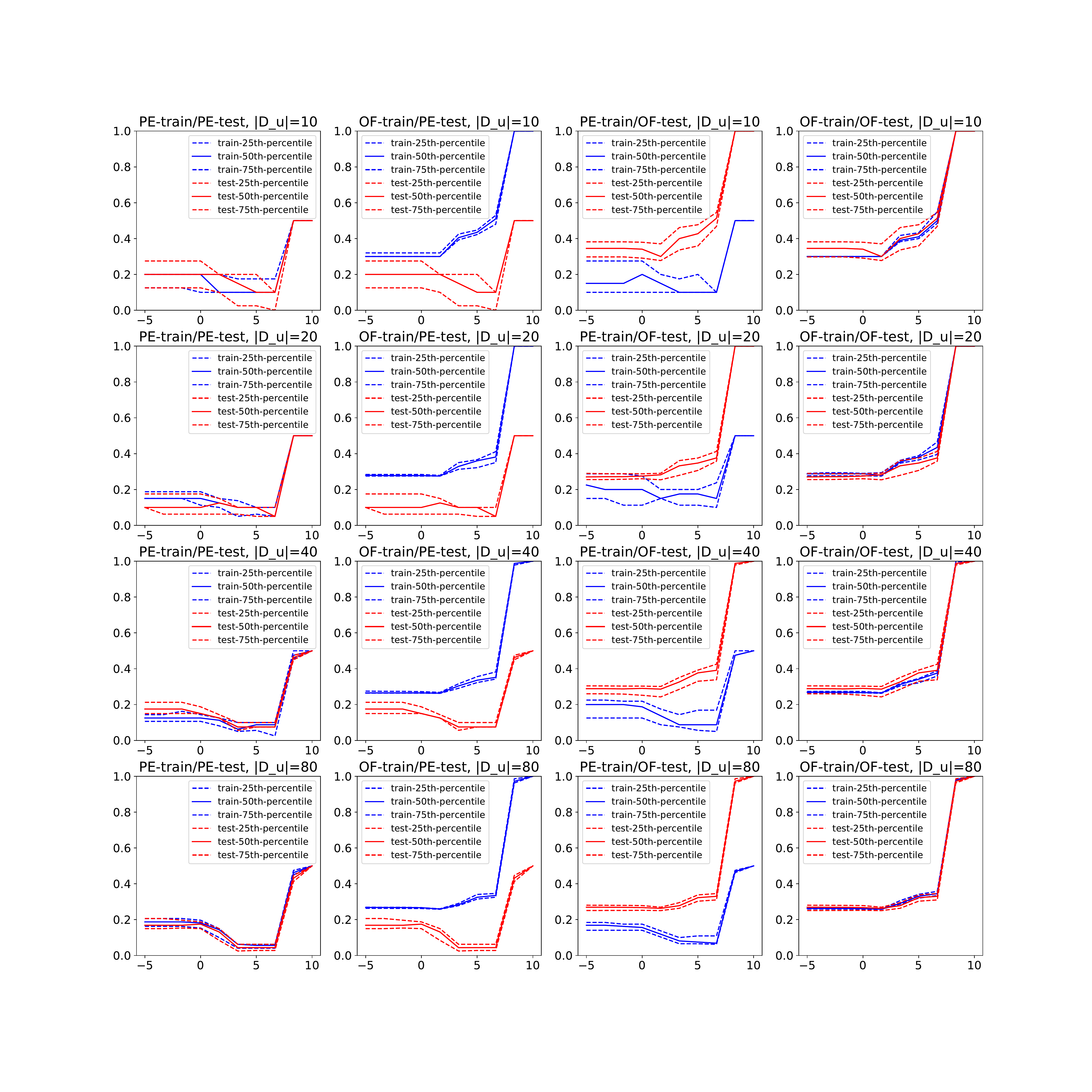}
\end{center}
\caption{Digit $k = \x$.
	Values of the training and testing objective functions 
	($\text{PE}$ or $\text{OF}$, $y$-axis) 
	against the logarithm of the model free parameter
	($\log \rho$, $x$-axis)
	for varying sizes of the datasets 
	$\lvert{\cal D}_u\rvert = 2 n_{\text{size}}$
	($u \in \{\text{pre-pre-train}, \text{pre-pre-test}, 
	\text{pre-test}, \text{test}\}$, 
	$n_{\text{size}} \in \{5, 10, 20, 40\}$).
  Solid and dashed lines represent the median and the 25th or 75th percentile of the values 
  obtained over $10$ equivalent experiments.
  As specified in the plot legends,
  blue and red lines are associated with training and testing objectives, respectively.
  Plots obtained for $k > 0$ are similar.}
\label{figure plots \x}
\end{figure}
}

\begin{figure}
\begin{center}
\includegraphics[trim=200 250 200 250, clip=1, width=.90\textwidth]{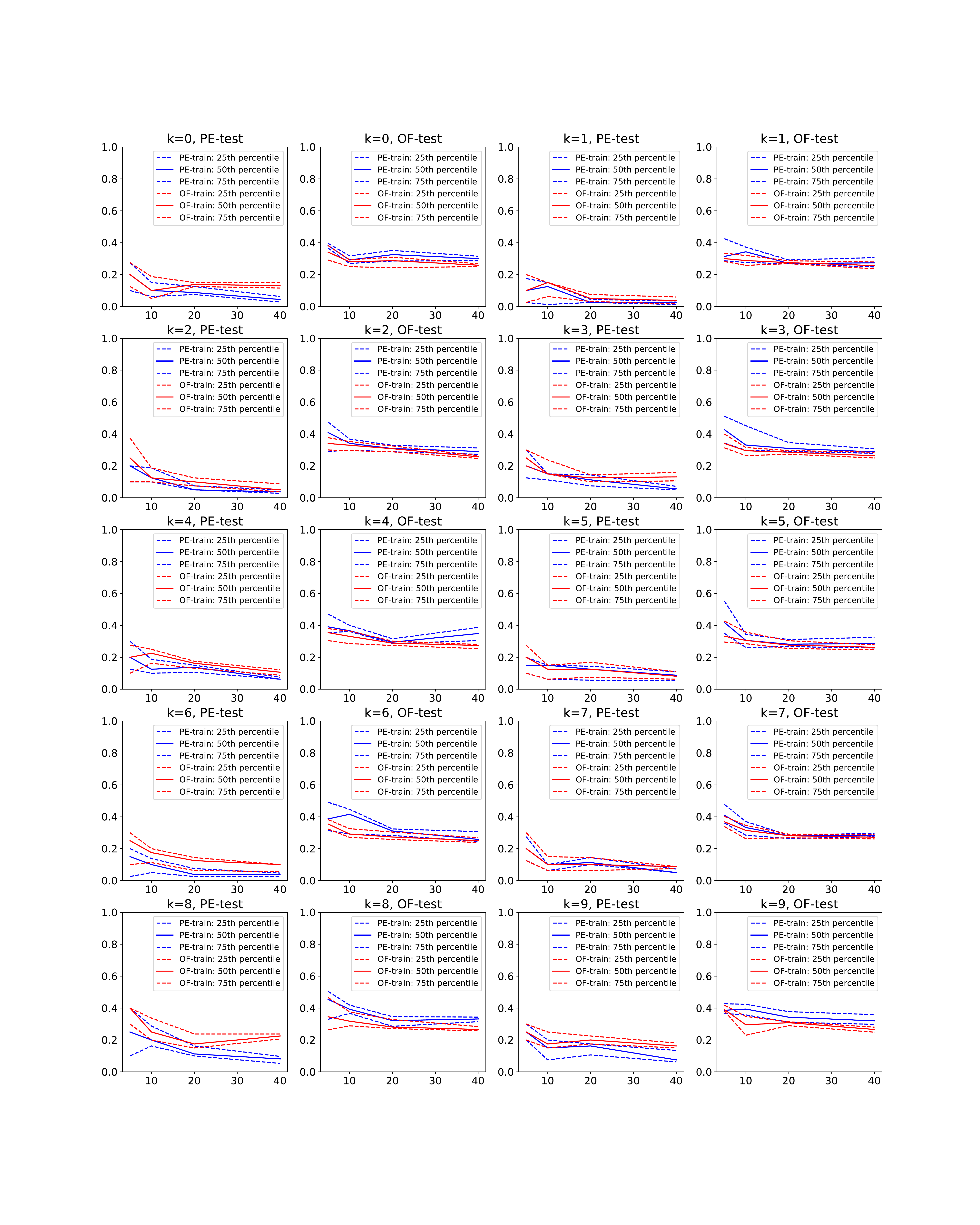}
\end{center}
\caption{\text{PE}- and \text{OF}-test scores against the size of the training and test sets
  $\lvert{\cal D}_u\rvert = 2 n_{\text{size}}$ ($u \in \{\text{pre-pre-train}, \text{pre-pre-test}, 
  \text{pre-test}, \text{test}\}$, 
	$n_{\text{size}} \in \{5, 10, 20, 40\}$).
	Plot titles specify the integer $k \in \{0, \dots, 9 \}$
	to be discriminated in the corresponding experiments 
	and the type of testing objective.
	Each plot shows the median (solid lines)
	and the 25th or 75th percentile of the values 
	obtained over $10$ equivalent experiments.
	Blue and red lines are associated with PE- and OF-training,
	respectively.}
\label{figure scores}
\end{figure}

\foreach \x in {5}
{
\begin{table}
\begin{center}
\begin{tabular}{|c|c|c|c|c|}
\hline
$k$&PE-test/PE-train&PE-test/OF-train&OF-test/PE-train&OF-test/OF-train\\
\hline
$0$&$0.190 \pm 0.145$&$0.230 \pm 0.155$&$0.402 \pm 0.108$&$0.336 \pm 0.052$\\
\hline
$1$&$0.120 \pm 0.108$&$0.130 \pm 0.110$&$0.367 \pm 0.113$&$0.321 \pm 0.060$\\
\hline
$2$&$0.180 \pm 0.108$&$0.220 \pm 0.154$&$0.386 \pm 0.101$&$0.351 \pm 0.066$\\
\hline
$3$&$0.210 \pm 0.083$&$0.250 \pm 0.112$&$0.432 \pm 0.117$&$0.361 \pm 0.075$\\
\hline
$4$&$0.210 \pm 0.114$&$0.190 \pm 0.114$&$0.410 \pm 0.085$&$0.342 \pm 0.070$\\
\hline
$5$&$0.150 \pm 0.081$&$0.180 \pm 0.098$&$0.445 \pm 0.098$&$0.359 \pm 0.103$\\
\hline
$6$&$0.150 \pm 0.128$&$0.200 \pm 0.134$&$0.407 \pm 0.116$&$0.353 \pm 0.047$\\
\hline
$7$&$0.200 \pm 0.110$&$0.210 \pm 0.114$&$0.426 \pm 0.086$&$0.367 \pm 0.051$\\
\hline
$8$&$0.260 \pm 0.150$&$0.360 \pm 0.102$&$0.417 \pm 0.100$&$0.363 \pm 0.107$\\
\hline
$9$&$0.230 \pm 0.110$&$0.260 \pm 0.150$&$0.449 \pm 0.187$&$0.412 \pm 0.089$\\
\hline
\end{tabular}
\end{center}
\caption{Case $n_{\text{size}} = \x$.
  The average and standard deviation of the scores obtained over 10 equivalent experiments
  with the sizes of the datasets $\lvert{\cal D}_u\rvert = 2 n_{\text{size}}$
  ($u \in \{ \text{pre-pre-train}, \text{pre-pre-test},	\text{pre-test}, \text{test}\}$.
  Integer $k \in \{0, \dots, 9\}$ in the first column is the digit to be discriminated in the corresponding experiments.}
\label{table size \x}
\end{table}
}

\foreach \x in {10}
{
\begin{table}
\begin{center}
\begin{tabular}{|c|c|c|c|c|}
\hline
$k$&PE-test/PE-train&PE-test/OF-train&OF-test/PE-train&OF-test/OF-train\\
\hline
$0$&$0.115 \pm 0.059$&$0.115 \pm 0.078$&$0.298 \pm 0.040$&$0.272 \pm 0.029$\\
\hline
$1$&$0.095 \pm 0.072$&$0.125 \pm 0.072$&$0.335 \pm 0.062$&$0.299 \pm 0.051$\\
\hline
$2$&$0.140 \pm 0.058$&$0.150 \pm 0.074$&$0.335 \pm 0.042$&$0.315 \pm 0.047$\\
\hline
$3$&$0.170 \pm 0.117$&$0.205 \pm 0.125$&$0.361 \pm 0.092$&$0.300 \pm 0.054$\\
\hline
$4$&$0.140 \pm 0.070$&$0.220 \pm 0.084$&$0.382 \pm 0.057$&$0.325 \pm 0.052$\\
\hline
$5$&$0.120 \pm 0.068$&$0.110 \pm 0.073$&$0.308 \pm 0.050$&$0.331 \pm 0.084$\\
\hline
$6$&$0.095 \pm 0.057$&$0.170 \pm 0.084$&$0.377 \pm 0.103$&$0.288 \pm 0.042$\\
\hline
$7$&$0.105 \pm 0.069$&$0.115 \pm 0.071$&$0.350 \pm 0.086$&$0.309 \pm 0.048$\\
\hline
$8$&$0.210 \pm 0.089$&$0.260 \pm 0.104$&$0.393 \pm 0.076$&$0.333 \pm 0.059$\\
\hline
$9$&$0.135 \pm 0.084$&$0.195 \pm 0.088$&$0.385 \pm 0.068$&$0.293 \pm 0.066$\\
\hline
\end{tabular}
\end{center}
\caption{Case $n_{\text{size}} = \x$.
  The average and standard deviation of the scores obtained over 10 equivalent experiments
  with the sizes of the datasets $\lvert{\cal D}_u\rvert = 2 n_{\text{size}}$
  ($u \in \{ \text{pre-pre-train}, \text{pre-pre-test},	\text{pre-test}, \text{test}\}$.
  Integer $k \in \{0, \dots, 9\}$ in the first column is the digit to be discriminated in the corresponding experiments.}
\label{table size \x}
\end{table}
}

\foreach \x in {20}
{
\begin{table}
\begin{center}
\begin{tabular}{|c|c|c|c|c|}
\hline
$k$&PE-test/PE-train&PE-test/OF-train&OF-test/PE-train&OF-test/OF-train\\
\hline
$0$&$0.097 \pm 0.044$&$0.140 \pm 0.032$&$0.320 \pm 0.049$&$0.276 \pm 0.037$\\
\hline
$1$&$0.035 \pm 0.030$&$0.058 \pm 0.043$&$0.280 \pm 0.021$&$0.276 \pm 0.020$\\
\hline
$2$&$0.063 \pm 0.017$&$0.102 \pm 0.039$&$0.310 \pm 0.027$&$0.306 \pm 0.024$\\
\hline
$3$&$0.125 \pm 0.065$&$0.125 \pm 0.034$&$0.316 \pm 0.041$&$0.280 \pm 0.020$\\
\hline
$4$&$0.130 \pm 0.035$&$0.153 \pm 0.048$&$0.303 \pm 0.032$&$0.287 \pm 0.022$\\
\hline
$5$&$0.115 \pm 0.070$&$0.123 \pm 0.070$&$0.291 \pm 0.055$&$0.274 \pm 0.038$\\
\hline
$6$&$0.053 \pm 0.045$&$0.130 \pm 0.076$&$0.307 \pm 0.036$&$0.280 \pm 0.028$\\
\hline
$7$&$0.120 \pm 0.037$&$0.102 \pm 0.061$&$0.280 \pm 0.027$&$0.277 \pm 0.022$\\
\hline
$8$&$0.125 \pm 0.042$&$0.183 \pm 0.058$&$0.325 \pm 0.051$&$0.299 \pm 0.041$\\
\hline
$9$&$0.143 \pm 0.043$&$0.205 \pm 0.063$&$0.343 \pm 0.038$&$0.303 \pm 0.021$\\
\hline
\end{tabular}
\end{center}
\caption{Case $n_{\text{size}} = \x$.
  The average and standard deviation of the scores obtained over 10 equivalent experiments
  with the sizes of the datasets $\lvert{\cal D}_u\rvert = 2 n_{\text{size}}$
  ($u \in \{ \text{pre-pre-train}, \text{pre-pre-test},	\text{pre-test}, \text{test}\}$.
  Integer $k \in \{0, \dots, 9\}$ in the first column is the digit to be discriminated in the corresponding experiments.}
\label{table size \x}
\end{table}
}
\foreach \x in {40}
{
\begin{table}
\begin{center}
\begin{tabular}{|c|c|c|c|c|}
\hline
$k$&PE-test/PE-train&PE-test/OF-train&OF-test/PE-train&OF-test/OF-train\\
\hline
$0$&$0.049 \pm 0.034$&$0.129 \pm 0.040$&$0.303 \pm 0.035$&$0.262 \pm 0.022$\\
\hline
$1$&$0.029 \pm 0.026$&$0.041 \pm 0.026$&$0.279 \pm 0.040$&$0.259 \pm 0.027$\\
\hline
$2$&$0.045 \pm 0.026$&$0.065 \pm 0.030$&$0.291 \pm 0.027$&$0.255 \pm 0.016$\\
\hline
$3$&$0.069 \pm 0.036$&$0.131 \pm 0.030$&$0.290 \pm 0.034$&$0.262 \pm 0.026$\\
\hline
$4$&$0.076 \pm 0.030$&$0.103 \pm 0.027$&$0.345 \pm 0.046$&$0.269 \pm 0.017$\\
\hline
$5$&$0.082 \pm 0.029$&$0.084 \pm 0.030$&$0.295 \pm 0.041$&$0.263 \pm 0.020$\\
\hline
$6$&$0.036 \pm 0.021$&$0.089 \pm 0.047$&$0.275 \pm 0.049$&$0.254 \pm 0.025$\\
\hline
$7$&$0.055 \pm 0.020$&$0.091 \pm 0.024$&$0.291 \pm 0.030$&$0.273 \pm 0.016$\\
\hline
$8$&$0.078 \pm 0.030$&$0.215 \pm 0.043$&$0.330 \pm 0.021$&$0.271 \pm 0.020$\\
\hline
$9$&$0.092 \pm 0.042$&$0.169 \pm 0.029$&$0.327 \pm 0.037$&$0.266 \pm 0.023$\\
\hline
\end{tabular}
\end{center}
\caption{Case $n_{\text{size}} = \x$.
  The average and standard deviation of the scores obtained over 10 equivalent experiments
  with the sizes of the datasets $\lvert{\cal D}_u\rvert = 2 n_{\text{size}}$
  ($u \in \{ \text{pre-pre-train}, \text{pre-pre-test},	\text{pre-test}, \text{test}\}$.
  Integer $k \in \{0, \dots, 9\}$ in the first column is the digit to be discriminated in the corresponding experiments.}
\label{table size \x}
\end{table}
}

We let $\rho \in {\cal R}$,
with ${\cal R} := \{e^{\min_\rho + (\max_\rho - \min_\rho) r / R} \}_{r = 0}^{R - 1}$,
$[\min_\rho, \max_\rho] := [-5, 10]$, $R := 10$.
For given integers $k \in \{0,\dots, 9\}$
and dataset sizes $n_{\text{size}} \in \{5, 10, 20, 40\}$,
we randomly extract from the MNIST database 10 quadruples of datasets
listed on the right-hand side of \eqref{eq:full-split} of equal sizes,
\begin{equation*}
  \left|
    {\cal D}_{\text{pre-pre-train}}
  \right|
  =
  \left|
  {\cal D}_{\text{pre-pre-test}}
  \right|
  =
  \left|
  {\cal D}_{\text{pre-test}}
  \right|
  =
  \left|
  {\cal D}_{\text{test}}
  \right|
  =
  2 n_{\text{size}}.
\end{equation*}
Each of the 400 datasets of size $2 n_{\text{size}}$
contains $n_{\text{size}}$ images of integer $k$ (with label $y = 1$)
and $n_{\text{size}}$ images of other integers $k' \ne k$ (with label $y = 0$).

All objects $x \in {\cal X}$ are vectorized $28\times 28$ grey-scale images of hand-written digits,
i.e., all $x = (x_1, \dots, x_{784})$ ($x_i \in [0, 1]$, $i = 1, \dots, 784$)
are such that the $(i, j)$th pixel of an image corresponds to the $(28(i - 1) + j)$th entry of the associated vector $x$.
The vectors are normalized to have the unit Euclidean length, $\| x \| = 1$.

For each of the 400 binary classification datasets
we run an independent training-testing experiment,
as explained in Section \ref{section methods model}.
Summaries of results are reported in Figures~\ref{figure plots 0}--\ref{figure scores}
and Tables~\ref{table size 5}--%
\ref{table size 40}.

We have already commented on the results in Tables~\ref{table size 5}--\ref{table size 40}:
when the goal is to design a good conformal predictor to be evaluated with the OF criterion,
OF training is preferable
(there is only one case in the tables where PE training works better);
and for the design of point classifiers to be evaluated using prediction accuracy,
PE training usually works better
(there are only 3 cases where OF training works better
and 2 cases where it works equally well).
This finding can be summarized by saying that \emph{consonant training}
(OF-training when the goal is OF-performance
and PE-training when the goal is PE-performance)
usually works better than \emph{dissonant training}
(OF-training when the goal is PE-performance
or PE-training when the goal is OF-performance).

Figure~\ref{figure plots 0} sheds light on the reasons for consonant training
working better than dissonant training.
In the case of consonant training (the first and fourth columns),
the performance curves for training and test sets look very similar,
and in many cases almost coincide.
In the case of dissonant training (the second and third columns),
they are not only at different levels
(which is to be expected since the PE and OF criteria produce numbers at different scales),
but their shapes look different, often attaining their minima at different places.

Figure~\ref{figure scores} gives, essentially, a different representation of the results
presented in Tables~\ref{table size 5}--\ref{table size 40}.
The test performance improves as the size of the training set grows,
albeit not very quickly.

\section{Conclusion}
\label{sec:conclusion}

In this paper we used a ``validation set'' (either the pre-test set or the pre-pre-test set)
for choosing the conformity measure to use at the testing stage.
There are ways to use the available training data more efficiently.
On the PE side, we can use cross-validation.
On the OF side, we can use cross-conformal prediction
(or the related procedure of jackknife+, \citealp{Barberetal2019})
instead of split-conformal prediction,
since the former are more economical with the data.
This is an interesting direction of further research.

The performance of conformal predictors can be further improved
(and provable validity of split-conformal prediction regained)
by using full conformal prediction.
However, in this case our methods will be computationally feasible
only when applied to a fairly narrow (but important) class of training procedures.

Other directions of further research include:
\begin{itemize}
\item
  Replacing exhaustive search over the discretized parameter space used in this paper
  by more efficient methods, such as Gradient Descent.
\item
  Experiments with other criteria of efficiency mentioned in \citet{vovk2016criteria},
  including those that are not probabilistic.
  (It is natural to expect that the prediction error for classifiers trained using such criteria suffers.)
\end{itemize}

\end{document}